\documentclass[utf8]{FrontiersinHarvard} 
\usepackage{url,hyperref,lineno,microtype,subcaption}
\usepackage{multirow,booktabs,tabularx}

\def\keyFont{\fontsize{8}{11}\helveticabold }
\def\firstAuthorLast{Liang {et~al.}} 
\def\Authors{Yueqing Liang\,$^{1}$, Canyu Chen\,$^{1}$, Tian Tian\,$^{2}$, and Kai Shu\,$^{1,*}$}

\def\Address{$^{1}$Department of Computer Science, Illinois Institute of Technology \\
$^{2}$Stuart School of Business, Illinois Institute of Technology}

\def\corrAuthor{10 W 31st St, Chicago, IL 60616}
\def\corrEmail{kshu@iit.edu}
 
\newcommand{\m}{\textsc{FairDA}}
\newtheorem{theorem}{Theorem}
\newcommand{\kai}[1]{\textcolor{black}{#1}}

\begin{document}
\onecolumn
\firstpage{1}

\title[{\m}]{Joint Adversarial Learning for Cross-domain Fair Classification} 
\title[Fair Classification via Domain Adaptation]{Fair Classification via Domain Adaptation: A Dual Adversarial Learning Approach}
\author[\firstAuthorLast ]{\Authors} 
\address{} 
\correspondance{} 

\extraAuth{}

\maketitle

\begin{abstract}
Modern machine learning (ML) models are becoming increasingly popular and are widely used in decision-making systems. However, studies have shown critical issues of ML discrimination and unfairness, which hinder their adoption on high-stake applications. 
Recent research on fair classifiers has drawn significant attention to developing effective algorithms to achieve fairness and good classification performance. Despite the great success of these fairness-aware machine learning models, most of the existing models require sensitive attributes to pre-process the data, regularize the model learning or post-process the prediction to have fair predictions. However, sensitive attributes are often incomplete or even unavailable due to privacy, legal or regulation restrictions. Though we lack the sensitive attribute for training a fair model in the target domain, there might exist a similar domain that has sensitive attributes. Thus, it is important to exploit auxiliary information from a similar domain to help improve fair classification in the target domain. Therefore, in this paper, we study a novel problem of exploring domain adaptation for fair classification. We propose a new framework that can learn to adapt the sensitive attributes from a source domain for fair classification in the target domain.  Extensive
experiments on real-world datasets illustrate the effectiveness of the proposed model for fair classification, even when no sensitive attributes are available in the target domain.
\tiny
 \keyFont{ \section{Keywords:} fair machine learning, adversarial learning, domain adaptation, trustworthiness, transfer learning} 
\end{abstract}

\section{Introduction}

The recent development of machine learning models has been increasingly used for high-stake decision making such as filtering loan applicants~\citep{hamid2016developing}, deploying police officers~\cite{mena2011machine}, etc. However, a prominent concern is when a learned model has bias against some specific demographic group such as race or gender. For example, a recent report shows that software used by schools to filter student applications may be biased toward a specific race group\footnote{https://www.fastcompany.com/90342596/schools-are-quietly-turning-to-ai-to-help-pick-who-gets-in-what-could-go-wrong}; and COMPAS~\cite{redmond2011communities}, a tool for crime prediction, is shown to be more likely to assign a higher risk score to African-American offenders than to Caucasians with the same profile. Bias in algorithms can emanate from unrepresentative or incomplete training data or flawed information that reflects historical inequalities~\cite{du2020fairness}, which can lead to unfair decisions that have a collective and disparate impact on certain groups of people. Therefore, it is important to ensure fairness in machine learning models.

\begin{figure}[ht!]
\begin{center}
\includegraphics[width=10cm]{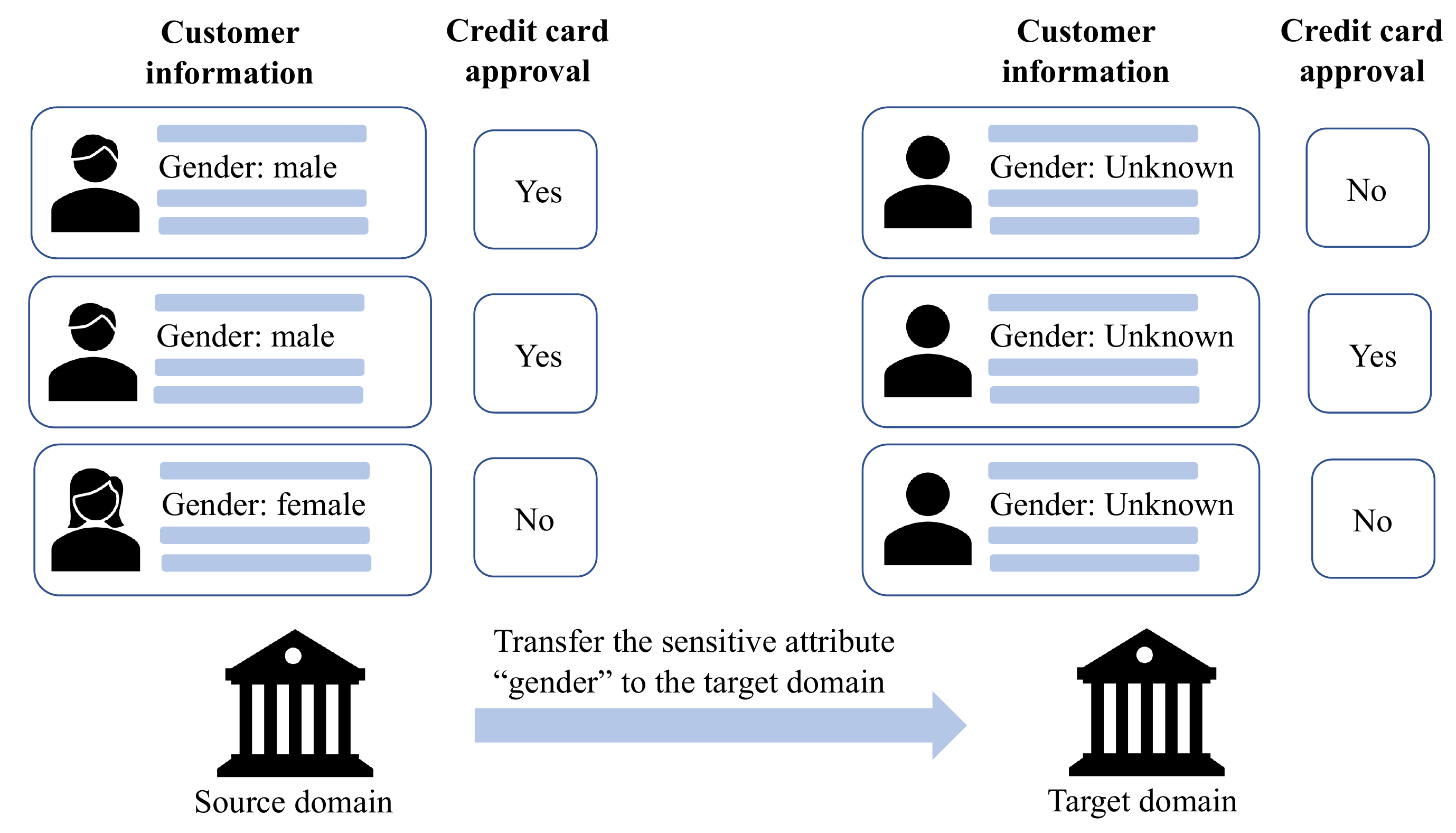}
\end{center}
\caption{An illustration of cross-domain fair classification. The information of the sensitive attribute is available in the source domain, while unavailable on the target domain. }\label{fig:scenario}
\end{figure}

Recent research on fairness machine learning has drawn significant attention to developing effective algorithms to achieve fairness and maintain good prediction performance~\cite{barocas2017fairness}. However, the majority of these models require sensitive attributes to preprocess the data, regularize the model learning or post-process the prediction to have fair predictions. For example, Kamiran \textit{et al.} propose to assign the weights of each training sample differently by reweighting to ensure fairness before model training~\cite{kamiran2012data}. In addition, Madras \textit{et al.} utilize adversarial training on sensitive attributes and prediction labels for debiasing classification results~\cite{madras2018learning}.
However, sensitive attributes are often incomplete or even unavailable due to privacy, legal, or regulation restrictions. For example, by the law of the US Consumer Financial Protection Bureau (CFPB), creditors may not request or collect information about an applicant’s race, color, sex, etc\footnote{https://www.consumerfinance.gov/rules-policy/regulations/1002/5/}.

Although we lack the sensitive attribute for training a fair model in the target domain, there might exist a similar domain that has sensitive attributes, which paves us a way to adapt the fairness from a related source domain to the target. For example, as shown in Fig.~\ref{fig:scenario}, when historical features and risk labels are available for credit risk assessment,
while user sensitive attributes (i.e., gender) are unavailable when expanding into new markets, we may learn models from its existing market data (source) to score people in the new market (target).

Though there are extensive works on domain adaptation, they are overwhelmingly on classification or knowledge transfer~\cite{zhuang2020comprehensive,tan2018survey}; while the work on fairness classification through domain adaptation is rather limited. 
 
Very few recent works study a \textit{shallow scenario of transfer} between different groups of sensitive attributes in the same domain for fair classification \cite{schumann2019transfer,coston2019fair}. We study a general cross-domain fair classification problem in terms of a specific sensitive attribute, which is not explored yet to our best knowledge.

Therefore, we propose a dual-adversarial learning framework to \textit{learn to adapt} sensitive attributes for fair classification in the target domain. In essence, we investigate the following challenges: (1) how to adapt the sensitive attributes to the target domain by transferring knowledge from the source domain; and (2) how to predict the labels accurately and satisfy fairness criteria. It is non-trivial since there are domain discrepancies across domains and no anchors are explicitly available for knowledge transfer. Our solutions to these challenges result in a novel framework called {\m} for fairness classification via domain adaptation. Our main contributions are summarized as follows:

\begin{itemize}
\item We study a novel problem of fair classification via domain adaptation.
\item We propose a new framework {\m} that simultaneously transfers knowledge to adapt sensitive attributes and learns a fair classifier in the target domain with a dual adversarial learning approach.  
\item We conduct theoretical analysis demonstrating that fairness in the target domain can be achieved with estimated sensitive attributes derived from the source domain.
\item We perform extensive experiments on real-world datasets to demonstrate the effectiveness of the proposed method for fair classification without sensitive attributes.
\end{itemize}

\section{Related Work}

\label{sec:related}
In this section, we briefly review the related works on fairness machine learning and deep domain adaptation.

\subsection{Fairness in Machine Learning}
Recent research on fairness in machine learning has drawn significant attention to developing effective algorithms to achieve fairness and maintain good prediction performance. which generally focus on individual fairness~\cite{kang2020inform} or group fairness~\cite{hardt2016equality,zhang2017achieving}. Individual fairness requires the model to give similar predictions to similar individuals~\cite{cheng2021socially,DBLP:journals/corr/abs-2010-04053}. In group fairness, similar predictions are desired among multiple groups categorized by a specific sensitive attribute (e.g., gender). Other niche notions of fairness include subgroup fairness~\cite{kearns2018preventing} and Max-Min fairness\cite{lahoti2020fairness}, which aims to maximize the minimum expected utility across groups. In this work, we focus on group fairness. 
To improve group fairness, debiasing techniques have been applied at different stages of a machine learning model: (1) \textit{Pre-processing} approaches~\cite{kamiran2012data} apply dedicated transformations on the original dataset to remove intrinsic discrimination and obtain unbiased training data prior to modeling; (2) \textit{In-processing} approaches~\cite{agarwal2018reductions,bechavod2017learning} tackle this issue by incorporating fairness constraints or fairness-related objective functions to the design of machine learning models; and (3) \textit{Post-processing} approaches~\cite{dwork2018decoupled} revise the biased prediction labels by debiasing mechanisms.

Although the aforementioned methods can improve group fairness, they generally require the access of sensitive attributes, which is often infeasible. Very few recent works study fairness with limited sensitive attributes. For example, \cite{zhao2021fair} and \cite{gupta2018proxy} explore and use the related features as proxies of the sensitive attribute to achieve better fairness results. \cite{lahoti2020fairness} proposes an adversarial reweighted method to achieve the Rawlsian Max-Min fairness objective which aims at improving the accuracy for the worst-case protected group. However, these methods may require domain knowledge to approximate sensitive attributes or are not suitable for ensuring group fairness. In addition, though ~\cite{madras2018learning} mentioned fairness transfer, what they study is the cross-task transfer~\cite{tan2021otce}, which is different from our setting.

In this paper, we study the novel problem of cross-domain fairness classification which aims to estimate the sensitive attributes of the target domain to achieve the group fairness. 

\subsection{Deep Domain Adaption}
Domain adaptation~\cite{pan2009survey} aims at mitigating the generalization bottleneck introduced by domain shift. With the rapid growth of deep neural networks, deep domain adaptation has drawn much attention lately. In general, deep domain adaptation methods aim to learn a domain-invariant feature space that can reduce the discrepancy between the source and target domains. This goal is accomplished either by transforming the features from one domain to be closer to the other domain, or by projecting both domains into a domain-invariant latent space~\cite{shu2019transferable}. For instance, TLDA~\cite{zhuang2015supervised} is a deep autoencoder-based model for learning domain-invariant representations for classification. Inspired by the idea of Generative Adversarial Network (GAN)~\cite{goodfellow2014generative}, researchers also propose to perform domain adaptation in an adversarial training paradigm~\cite{ganin2016domain,tzeng2017adversarial,shu2019transferable}. By exploiting a domain discriminator to distinguish the domain labels while learning deep features to confuse the discriminator, DANN~\cite{ganin2016domain} achieves superior domain adaptation performance. ADDA~\cite{tzeng2017adversarial} learns a discriminative representation using labeled source data and then maps the target data to the same space through an adversarial loss. Recently, very few works apply transfer learning techniques for fair classification \cite{schumann2019transfer,coston2019fair}. Although sensitive attributes are not available in the target domain, there may exist some publicly available datasets which can be used as auxiliary sources. However, these methods mostly consider a shallow scenario of transfer between different sensitive attributes in the same dataset.

In this paper, we propose a new domain adaptive approach based on dual adversarial learning to achieve fair classification in the target domain.

\section{Problem Statement}\label{sec:problem}

We first introduce the notations of this paper, and then give the
formal problem definition. Let $\mathcal{D}^1=\{\mathcal{X}^1,\mathcal{A}^1, \mathcal{Y}^1\}$ denote the data in the source domain, where $\mathcal{X}^1$, $\mathcal{A}^1$, and $\mathcal{Y}^1$ represent the set of data samples, sensitive attributes, and corresponding labels.
Let $\mathcal{D}^2=\{\mathcal{X}^2,\mathcal{Y}^2\}$ be the target domain, where the sensitive attributes of the target domain are unknown. The data distributions in the source and target domains are similar but different with the domain discrepancy. Following existing work on fair classification ~\cite{barocas2017fairness,mehrabi2019survey}, we evaluated fairness performance using metrics such as equal opportunity and demographic parity. Without loss of generality, we consider binary classification. Equal opportunity requires that the probability of positive instances with arbitrary sensitive attributes $A$ being assigned to a positive outcome are equal: ${\mathbb{E}}(\hat{Y} \mid A=a, Y=1) = {\mathbb{E}}(\hat{Y} \mid A=b, Y=1)$, where $\hat{Y}$ is predicted label. Demographic parity requires the behavior of the prediction model to be fair to different sensitive groups. Concretely, it requires that the positive rate across sensitive attributes are equal: ${\mathbb{E}}(\hat{Y} \mid A=a) = {\mathbb{E}}(\hat{Y} \mid A=b), \forall a, b$. The problem of fair classification with domain adaptation is formally defined as follows:

\begin{center}
\fbox{\parbox[c]{.95\linewidth}{\textbf{Problem Statement:}
Given the training data $\mathcal{D}^1$ and $\mathcal{D}^2$ from the source and target domain, learn an effective classifier for the target domain while satisfying the fairness criteria such as demographic parity.}}
\end{center}

\section{{\m}: Fair Classification with Domain Adaptation}\label{sec:model}

\begin{figure}[t!]
\begin{center}
\includegraphics[width=0.6\textwidth]{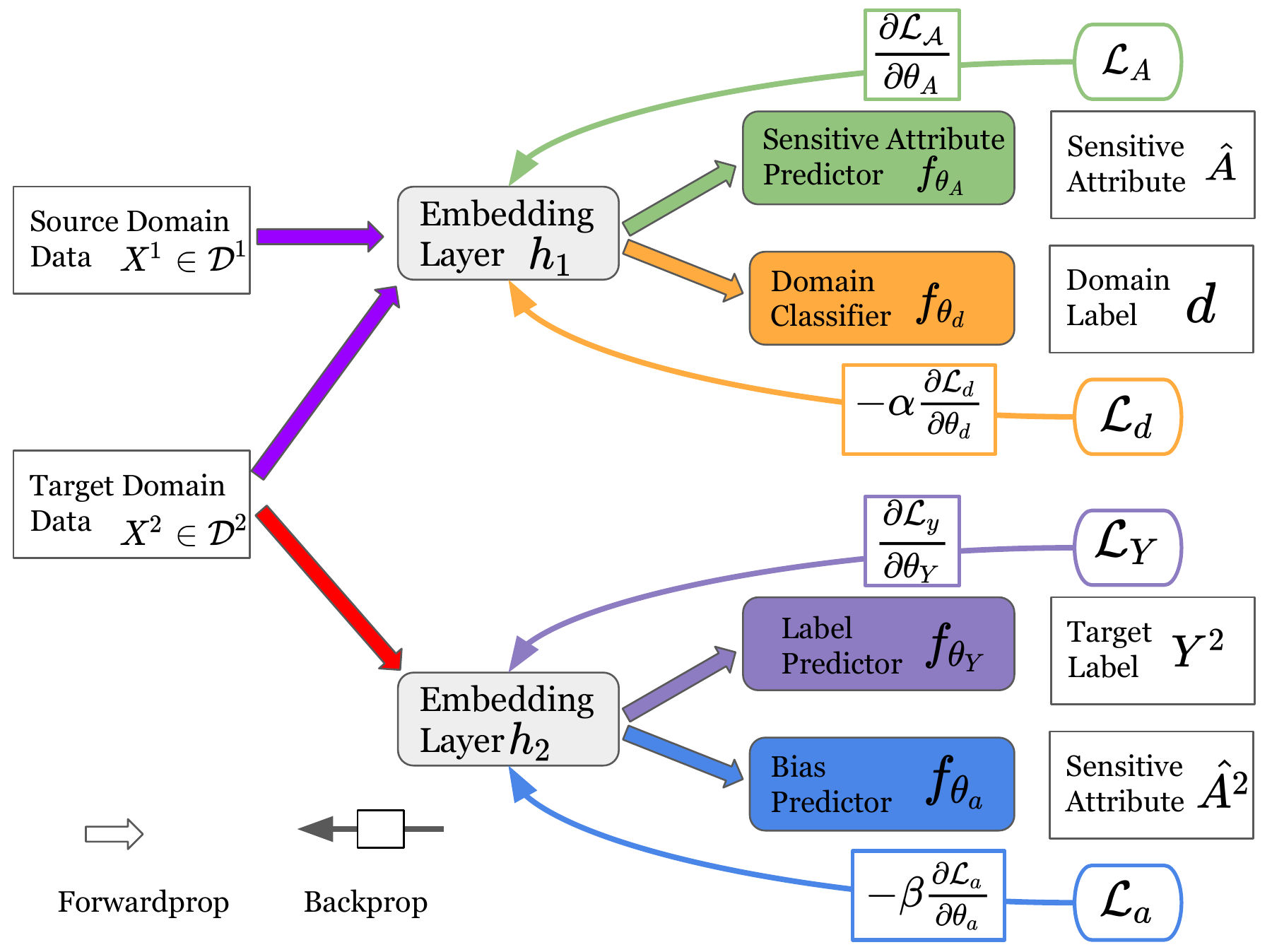}
\end{center}
\caption{An illustration of the proposed framework {\m}. It consists of two major modules: (1) an adversarial domain adaptation module containing a sensitive attribute predictor and a domain classifier; and (2) an adversarial debiasing module consisting of a label predictor and a bias predictor.}\label{fig:framework}
\end{figure}

In this section, we present the details of the proposed framework for fair classification with domain adaptation. As shown in Figure~\ref{fig:framework}, our framework consists of two major modules: (1) an adversarial domain adaptation module that estimates sensitive attributes for the target domain; and (2) an adversarial debiasing module to learn a fair classifier in the target domain. 

Specifically, first, the adversarial domain adaptation module contains a sensitive attribute predictor $f_{\theta_A}$ that describes the modeling of predicting sensitive attributes and a domain classifier $f_{\theta_d}$ that illustrates the process of transferring knowledge to the target domain for estimating sensitive attributes. In addition, the adversarial debiasing module consists of a label predictor $f_{\theta_Y}$ that models the label classification in the target domain, and a bias predictor $f_{\theta_a}$ to illustrate the process of differentiating the estimated sensitive attributes from the target domain data.

\subsection{Estimating Target Sensitive Attributes}

Since the sensitive attributes in the target domain are unknown, it is necessary to estimate the target sensitive attributes to build fair classifiers. Recent advancement in unsupervised domain adaptation has shown promising results to adapt a classifier on a source domain to the unlabeled data to the target domain~\cite{ganin2015unsupervised}. Therefore, we propose to investigate an unsupervised domain adaptation framework to infer the sensitive attributes in the target domain. Specifically, to minimize the prediction error of sensitive attributes in the source domain, with the following objective function:
 \begin{align}
&\min_{\theta_{A}}\mathcal{L}_{A} =  \mathbb{E}_{(X^1,A^1)\in \mathcal{D}^1}\ell(A^1, f_{\theta_A}(h_1(X^1)))
 \end{align}
where $f_{\theta_{A}}$ is used to predict sensitive attributes, $\ell$ denotes the loss function to minimize prediction error, such as cross-entropy loss, and the embedding function $h_1(\cdot)$ is to learn the representation for predicting sensitive attributes. To ensure that $f_{\theta_{A}}$ trained on $\mathcal{D}^1$ can well estimate the sensitive attributes of $\mathcal{D}^2$. We further introduce a domain classifier $f_{\theta_d}$ to make the representation of source and target domains in the same feature space. Specifically,  $f_{\theta_d}$ aims to differentiate if a representation is from the source or the target; while $h_1(\cdot)$ aim to learn a domain-invariant representation that can fool $f_{\theta_d}$, i.e.,
\begin{align}
    \min_{\theta_{h_1}}\max_{\theta_d}\mathcal{L}_{d}= & \mathbb{E}_{X^1\in \mathcal{D}^1}[\log (f_{\theta_d}(h_1(X^1))] + \\ &\mathbb{E}_{X^2\in \mathcal{D}^2}[\log (1-(f_{\theta_d}(h_1(X^2)))] \nonumber
\end{align} 
where $\theta_d$ are the parameters of the domain classifier. The overall objective function for this module is a min-max game between the sensitive attributes predictor $f_{\theta_A}$ and the domain classifier $f_{\theta_d}$ as follows: 
\begin{align}\label{eqn:l1}
\min_{\theta_{A},\theta_{h_1}}\max_{\theta_d}\mathcal{L}_1 = \mathcal{L}_{A}-\alpha \mathcal{L}_{d}
\end{align}
where $\alpha$ controls the importance of the adversarial learning of the domain classifier. To estimate the sensitive attributes in the target domain, we will utilize the following function: $\hat{A}^2=f_{\theta_{A}}(h_1(X^2))$.

\subsection{Adversarial Debiasing for Fair Classification}

To learn a fair classifier for the target domain, we will leverage the derived sensitive attributes $\hat{A}^2$ and the labels $Y^2$ in the target domain. Specifically, we aim to learn a representation that can predict the labels accurately, while being irrelevant to sensitive attributes. To this end, we propose to feed the representation into a label predictor and a bias predictor. Specifically, the label predictor is to minimize the predicted errors of the labels with the following objective function:
 \begin{align}
\min_{\theta_{Y}}\mathcal{L}_{Y} =  \mathbb{E}_{(X^2,Y^2)\in \mathcal{D}^2}\ell(Y^2, f_{\theta_Y}(h_2(X^2)))
 \end{align}
where $f_{\theta_Y}$ is to predict the labels,  $h_2(\cdot)$ is an embedding layer to encode the features into a latent representation space, and $\ell$ is a cross-entropy loss. Furthermore, to learn fair representations and make fair predictions in the target domain, we incorporate a bias predictor to predict the sensitive attributes; while $h_2(\cdot)$ trying to learn the representation that can fool $f_{\theta_Y}$:
\begin{align}
\min_{\theta_{h_2}}\max_{\theta_a}\mathcal{L}_a=& \mathbb{E}_{X^2 \sim p(X^2 \mid \hat{A}^2=1)}[\log (f_{\theta_a}(h_2(X^2)))]+ \\&
\mathbb{E}_{X^2\sim p(X^2 \mid \hat{A}^2=0)}[\log (1-f_{\theta_a}(h_2(X^2)))] \nonumber
\end{align}
where $\theta_a$ are the parameters for the adversary to predict sensitive attributes. Finally, the overall objective function of adversarial debiasing for fair classification is a min-max function:
\begin{align}\label{eqn:l2}
\min_{\theta_{h_2},\theta_Y}\max_{\theta_a}\mathcal{L}_2 = \mathcal{L}_Y-\beta \mathcal{L}_a
\end{align}
where $\beta$ controls the importance of the bias predictor.

\subsection{The Proposed Framework: {\m}}
We have introduced how we can estimate sensitive attributes by adversarial domain adaptation, and how to ensure fair classification with adversarial debiasing. We integrate the components together and the overall objective function of the \textit{dual} adversarial learning is as follows:
\begin{align}
{\min}_{\theta_{h_1},\theta_{h_2},\theta_Y,\theta_A}{\max}_{\theta_d,\theta_a}\mathcal{L} = \mathcal{L}_1+\mathcal{L}_2
\end{align}
The parameters in the objective are learned through RMSProp, which is an adaptive learning rate method that divides the learning rate by an exponentially decaying average of squared gradients. We choose RMSProp as the optimizer because it is a popular and effective method to determine the learning rate abortively which is widely used for training adversarial neural networks~\cite{zhou2022flooddan, dou2019pnp, li2022unsupervised}. Adam is another widely adopted optimizer that extends RMSProp with momentum terms, however, the momentum terms may make Adam unstable\cite{mao2017least, luo2018wgan,clavijo2021adversarial}. We will also prioritize the training of $f_{\theta_A}$ to ensure a good estimation of $\hat{A}^2$. Next, we will conduct a theoretical analysis of the dual adversarial learning for fairness guarantee.

\section{A Theoretical Analysis on Fairness  Guarantee}
In this section, we perform a theoretical analysis of the fairness guarantee under the proposed framework {\m} with key assumptions. The model essentially contains two modules: (1) the adversarial domain adaptation for estimating sensitive attributes ($\mathcal{L}_1$ in Eqn.~\ref{eqn:l1}); and (2) the adversarial debiasing for learning a fair classifier ($\mathcal{L}_2$ in Eqn.~\ref{eqn:l2}). The performance of the second module is relying on the output (i.e., $\hat{A}^2$) of the first module.

To understand the theoretical guarantee for the first module, we follow the recent work on analyzing the conventional unsupervised domain adaptation~\cite{liu2021adversarial,zhang2019bridging}. The majority of them consider the model effectiveness under different types of domain shifts such as label shift, covariate shift, conditional shift, etc~\cite{liu2021adversarial,kouw2018introduction}. When analyzing domain adaptation models, researchers usually consider the existence of one shift while assuming other shifts are invariant across domains \cite{kouw2018introduction}. The existence of label shift is a common setting of unsupervised domain adaptation \cite{azizzadenesheli2019regularized, chen2018re, lipton2018detecting, wu2019domain}. We follow the above papers to make the same assumption on label shift, which in our scenario is \textit{sensitive attribute shift} across two domains, i.e., the prior distribution changes as $p(A^1)\neq p(A^2)$. In addition, we assume that the other shifts are invariant across domains, for example, $p(X^1|A^1)=p(X^2|A^2)$. Under the assumption of the sensitive attribute shift, we can derive the risk in the target domain as~\cite{kouw2018introduction}:
\begin{align}
    R^2 (h_1,\theta_A)= &\sum_{A\in\mathcal{A}} \int_{\mathcal{X}} \ell(A, f_{\theta_A}(h_1(X)))\frac{p(X^2|A^2)p(A^2)}{p(X^1|A^1)p(A^1)}p(X^1,A^1)\mathrm{d} X
  \\   =& \sum_{A\in\mathcal{A}} \int_{\mathcal{X}} \ell(A, f_{\theta_A}(h_1(X)))\frac{p(A^2)}{p(A^1)}p(X^1,A^1)\mathrm{d} X \nonumber
\end{align}
where the ratio $p(A^2)/p(A^1)$ represents the change in the proportions in the sensitive attribute. Since we do not have samples with sensitive attributes from the target domain, we can use the samples drawn from the source distribution to estimate the target sensitive attributes distribution with mean matching~\cite{gretton2009covariate} by minimizing the following function $\|M^1 p(A^1)-\mu^2\|_2^2$:
where $M^1$ is the vector empirical sample means from the source domain, i.e., $[\mu^1(f_{\theta_A}(h_1(X^1))|A^1=0),\mu^1(f_{\theta_A}(h_1(X^1))|A^1=1)]$. $\mu^2$ is the encoded feature means for the target. We will incorporate the above strategy for estimating the target sensitive attribute proportions with gradient descent during the adversarial training~\cite{li2019target}.

The noise induced for estimating the target sensitive attributes $\hat{A}^2$ is nonnegligible, which may influence the adversarial debiasing in the second module. Next, we theoretically show that under mild conditions, we can satisfy fairness metrics such as demographic parity. First, we can prove that the global optimum of $\mathcal{L}_2$ can be achieved if and only if\\ \indent \indent \indent \indent \indent \indent $p(\kai{Z^2}|\hat{A}^2=1)=p(\kai{Z^2}|\hat{A}^2=0)$, \kai{where $Z^2=h_2(X^2)$ is the representation of $X^2$}\\ according to the Proposition 1. in~\cite{goodfellow2014generative}.  Next, we introduce the following theorem under two reasonable assumptions:
\noindent \begin{theorem}\label{theo1}
Let $\hat{Y}^2$ denote the predicted label of the target domain, if\\
(1) The estimated sensitive attributes $\hat{A}^2$ and \kai{the representation of} $X^2$ are independent conditioned on the true sensitive attributes, i.e.,\\
\indent \indent \indent \indent \indent \indent $p(\hat{A}^2, \kai{Z^2}|A^2)=p(\hat{A}^2|A^2)p(\kai{Z^2}|A^2)$;\\
(2) The estimated sensitive attributes are not random, i.e.,\\
\indent \indent \indent \indent \indent \indent $p(A^2=1|\hat{A}^2=1)\neq p(|A^2=0|\hat{A}^2=1)$.\\
If $\mathcal{L}_2$ reaches the global optimum, the label prediction $f_{\theta_Y}$ will achieve demographic parity, i.e.,\\
\indent \indent \indent \indent \indent \indent $p(\hat{Y}^2|A^2=0)=p(\hat{Y}^2|A^2=1)$.
\end{theorem}
\noindent We first explain the two assumptions:\\
(1) since we use two separate embedding layers $h_1(\cdot)$ and $h_2(\cdot)$ to predict the target sensitive attributes, and learn the latent presentation, it generally holds that $\hat{A}^2$ is independent of the representation of $X^2$, i.e., $p(\hat{A}^2, \kai{Z^2}|A^2)=p(\hat{A}^2|A^2)p(\kai{Z^2}|A^2)$;\\ (2) Since we are using adversarial learning to learn an effective estimator $f_{\theta_A}$ for sensitive attributes, it is reasonable to assume that it does not produce random prediction results. \\
We prove Theorem~\ref{theo1} as follows:\\
Since $p(\hat{A}^2, \kai{Z^2}|A^2)=p(\hat{A}^2|A^2)p(\kai{Z^2}|A^2)$, we have $p(\kai{Z^2}|\hat{A}^2,A^2)=p(\kai{Z^2}|A^2)$. In addition, when the algorithm converges, we have $p(\kai{Z^2}|\hat{A}^2=1)=p(\kai{Z^2}|\hat{A}^2=0)$, which is equivalent with\\
$\sum_{A^2}p(\kai{Z^2}, A^2|\hat{A}^2=1)=\sum_{A^2}p(\kai{Z^2},A^2|\hat{A}^2=0)$. Therefore,
\begin{align}
\sum_{A^2}p(\kai{Z^2}|A^2)p(A^2|\hat{A}^2=1) = \sum_{A^2}p(\kai{Z^2}|A^2)p(A^2|\hat{A}^2=0)
\end{align}
Based on the above equation, we can get,
\begin{align}
    \frac{p(\kai{Z^2}|A^2=1)}{p(\kai{Z^2}|A^2=0)}=\frac{p(A^2=0|\hat{A}^2=1)-p(A^2=0|\hat{A}^2=0)}{p(A^2=1|\hat{A}^2=0)-p(A^2=1|\hat{A}^2=1)}=1
\end{align}
which shows that a global minimum achieves, i.e., $p(\kai{Z^2}|A^2=1)=p(\kai{Z^2}|A^2=0)$. Since $\hat{Y}^2=f_{\theta_Y}(\kai{Z^2})$, we can get $p(\hat{Y}^2|A^2=1)=p(\hat{Y}^2|A^2=0)$, which is the demographic parity.

\section{Experiments}\label{sec:experiments}
In this section, we present the experiments to evaluate the effectiveness of {\m}. We aim to answer the following research questions~(RQs):
\begin{itemize}
    \item \textbf{RQ1}: Can {\m} obtain fair predictions without accessing sensitive attributes in the target domain? 
    \item \textbf{RQ2}: How can we transfer fairness knowledge from the source domain while effectively using it to regularize the target's prediction? 
    \item \textbf{RQ3}: How would different choices of the weights of the two adversarial components impact the performance of {\m}?
\end{itemize}

\subsection{Datasets}\label{sec:datasets}
We conduct experiments on four publicly available benchmark datasets for fair classification: COMPAS~\cite{Julia2016machine}, Adult~\cite{asuncion2007uci}, Toxicity \cite{dixon2018measuring}
 and CelebA \cite{liu2018large}. 

\begin{itemize}
    \item \textbf{COMPAS}\footnote{\href{https://github.com/propublica/compas-analysis}{https://github.com/propublica/compas-analysis}}: This dataset describes the task of predicting the recidivism of individuals in the U.S. Both ``sex" and ``race" could be the sensitive attributes of this dataset \cite{le2022survey}. 
    \item \textbf{ADULT}\footnote{\href{https://archive.ics.uci.edu/ml/machine-learning-databases/adult/}{https://archive.ics.uci.edu/ml/machine-learning-databases/adult/}}: This dataset contains records of personal yearly income, and the label is whether the income of a specific individual exceeds 50k or not. Following \cite{zhao2021fair}, we choose ``sex" as the sensitive attribute.
    \item \textbf{Toxicity}\footnote{\href{https://www.kaggle.com/c/jigsaw-unintended-bias-in-toxicity-classification}{https://www.kaggle.com/c/jigsaw-unintended-bias-in-toxicity-classification}}: This dataset collects comments with labels indicating whether each comment is toxic or not. Following \cite{chuang2021fair}, we choose ``race" as the sensitive attribute.
    \item \textbf{CelebA}\footnote{\href{http://mmlab.ie.cuhk.edu.hk/projects/CelebA.html?fbclid=IwAR1VeZyPhkTzsoD_Fq8ItPwvyA0W1MD7fHO0v7MVaps1oX1fSt95q5i8Wfo}{http://mmlab.ie.cuhk.edu.hk/projects/CelebA.html}}: This dataset contains celebrity images, where each image has 40 human-labeled binary attributes. Following~\cite{chuang2021fair}, we choose the attribute ``attractive" as our prediction task and ``sex" as the sensitive attribute. 
\end{itemize}

The evaluation of all models is measured by the performance on the target dataset. We split train : eval : test set for the target dataset as 0.5 : 0.25 : 0.25. For the source dataset, because we do not evaluate on it, we split train : eval set as 0.6 : 0.4. It is worth noting that for the aforementioned datasets, we need to define a filter to derive source and target domains for empirical studies. Our general principle is to choose a feature that is reasonable in the specific scenario. For example, in COMPAS, we use \texttt{age} as the filter due to the prior knowledge that individuals in different age groups have different preferences to disclose their sex and race information. For ADULT, we consider the sex of each individual to be the sensitive attribute, and define two filters based on the \texttt{working class} and \texttt{country} to derive the source and target domains. The statistics of the datasets are shown in Table~\ref{tab:data}. 

\begingroup
\renewcommand{\arraystretch}{1.3}
\begin{table*}
    \centering
    \caption{Statistics of datasets. \textit{SA} refers to sensitive attributes. The filter defines the rule to group the original data to the datasets of the source and target domains. }
    \begin{tabular}{lcccccccc}
    \toprule
        Exp. & Ori. data & SA & Filter  & Dataset & Domain & SA avail. & \# Instance\\ \hline
        \multirow{2}{*}{1} & \multirow{2}{*}{COMPAS} & \multirow{2}{*}{Sex}& age $<$ 24  & COMPAS 1 & Source & Yes & 881 \\ 
        \cline{4-8}
        ~ & ~ &  & age $\geq$ 24 & COMPAS 2 & Target & No & 4,397 \\ \hline
        
        \multirow{2}{*}{2} & \multirow{2}{*}{COMPAS} & \multirow{2}{*}{Race}& age $<$ 24  & COMPAS 3 &Source & Yes & 881 \\ 
        \cline{4-8}
        ~ & ~ &  & age $\geq$ 24 & COMPAS 4 & Target & No & 4,397 \\ \hline
        
        \multirow{2}{*}{3} & \multirow{2}{*}{ADULT} & \multirow{2}{*}{Sex}& work\_class $\neq$ private  & ADULT 1 &Source & Yes & 11,915 \\ 
        \cline{4-8}
        ~ & ~ &  & work\_class $=$ private & ADULT 2 & Target & No & 33,307 \\ \hline
       \multirow{2}{*}{4} & \multirow{2}{*}{ADULT} & \multirow{2}{*}{Sex}& country $=$ US  & ADULT 3 &Source & Yes & 41,292 \\ 
        \cline{4-8}
        ~ & ~ &  & country $\neq$ US & ADULT 4 & Target & No & 3,930 \\ \hline
        
        \multirow{2}{*}{5} & \multirow{2}{*}{Toxicity} & \multirow{2}{*}{Race}& Created date  & Toxic 1 &Source & Yes & 11,010 \\ 
        \cline{4-8}
        ~ & ~ &  & Created date & Toxic 2 & Target & No & 7,340 \\ \hline

        \multirow{2}{*}{6} & \multirow{2}{*}{CelebA} & \multirow{2}{*}{Sex}& Young  & CelebA 1 &Source & Yes & 121,560 \\ 
        \cline{4-8}
        ~ & ~ &  & Young & CelebA 2 & Target & No & 81,040 \\ \bottomrule
        
    \end{tabular}\label{tab:data}
    
\end{table*}
\endgroup

\subsection{Experimental Settings}\label{sec:setting}

\subsubsection{Baselines}
Since there is no existing work on cross-domain fair classification, we compare our proposed {\m} with the following representative methods in fair classification without sensitive attributes. 
\begin{itemize}

    \item \textbf{Vanilla}: This method directly trains a base classifier without explicitly regularizing on the sensitive attributes.

    \item \textbf{ARL}~\cite{lahoti2020fairness}: This method utilizes reweighting on under-represented regions detected by adversarial model to alleviate bias. It focuses on improving Max-Min fairness rather than group fairness in our setting. However, it is an important work in fair machine learning. For the integrity of our experiment, we still involve it as one of our baselines. 
    \item \textbf{KSMOTE}~\cite{yan2020fair}: It first derives pseudo groups, and then uses them to design regularization to ensure fairness.
    \item \textbf{FairRF}~\cite{zhao2021fair}: It optimizes the prediction fairness without sensitive attribute but with some available features that are known to be correlated with the sensitive attribute. 
\end{itemize}

For KSMOTE\footnote{\href{https://imbalanced-learn.org/stable/}{https://imbalanced-learn.org/stable/}} and FairRF\footnote{\href{https://github.com/TianxiangZhao/fairlearn}{https://github.com/TianxiangZhao/fairlearn}}, we directly use the code provided by the authors. For all other approaches, we adopt a three-layer multi-layer-perceptron (MLP) as an example of the base classifier, and set the two hidden dimensions of MLP as 64 and 32, and use RMSProp optimizer with 0.001 as the initial learning rate. We have similar observations for other types of classifiers such as Logistic Regression and SVM.

\subsubsection{Evaluation Metrics}
Following existing work on fair models, we measure the classification performance with Accuracy \textbf{(ACC)} and \textbf{F1}, and the fairness performance based on \textbf{Demographic Parity} and \textbf{Equal Opportunity}~\cite{mehrabi2019survey}. We consider the scenario when the sensitive attributes and labels are binary, which can be naturally extended to more general cases.

\begin{itemize}
    \item Demographic Parity: A classifier is considered to be fair if the prediction $\hat{Y}$ is independent of the sensitive attribute $A$. In other words, demographic parity requires each demographic group has the same chance for a positive outcome: 
$\mathbb{E}(\hat{Y}|A=1)=\mathbb{E}(\hat{Y}|A=0)$. 
We will report the difference of each sensitive group's demographic parity ($\Delta_{DP}$):
\begin{align}
    \Delta_{DP}=|\mathbb{E}(\hat{Y}|A=1)-\mathbb{E}(\hat{Y}|A=0)|
\end{align}

\item Equal Opportunity: Equal opportunity considers one more condition than demographic parity. A classifier is said to be fair if the prediction $\hat{Y}$ of positive instances is independent of the sensitive attributes. Concretely, it requires the true positive rates of different sensitive groups to be equal: $\mathbb{E}(\hat{Y}|A=1, Y=1)=\mathbb{E}(\hat{Y}|A=0, Y=1)$. Similarly, in the experiments, we report the difference of each sensitive group's equal opportunity ($\Delta_{EO}$):
\begin{align}
    \Delta_{EO}=|\mathbb{E}(\hat{Y}|A=1, Y=1)-\mathbb{E}(\hat{Y}|A=0, Y=1)|
\end{align}
\end{itemize}
Note that demographic parity and equal opportunity measure fairness performance in different ways, and the smaller the values are, the better the performance of fairness.

\begingroup
\renewcommand{\arraystretch}{1.3}

\begin{table*}
\centering \caption{Fairness performance comparison in the COMPAS dataset. }
\begin{tabularx}{0.9\linewidth}{ccl|llll}
\toprule
\multicolumn{1}{l}{Datasets} & \multicolumn{1}{l}{Exp.} & Methods           & ACC         & F1          & $\Delta_{DP}$          & $\Delta_{EO}$          \\ \hline
\multirow{12}{*}{{COMPAS}}     & \multirow{6}{*}{1}       & Vanilla      & 0.677$\pm$0.015 & 0.639$\pm$0.018 & 0.128$\pm$0.015 & 0.137$\pm$0.005 \\ \cline{3-7}
                             &                          & ARL      & 0.672$\pm$0.008 & 0.609$\pm$0.032 & 0.109$\pm$0.038 & 0.119$\pm$0.057 \\
                             &                          & KSMOTE      & 0.632$\pm$0.012 & 0.525$\pm$0.018 & 0.136$\pm$0.017 & 0.146$\pm$0.018 \\
                             &                          & FairRF & 0.641$\pm$0.013 & 0.527$\pm$0.064 & 0.086$\pm$0.020  & 0.108$\pm$0.048 \\ \cline{3-7}
                             &                          & {\m}            & 0.661$\pm$0.023 & 0.639$\pm$0.016 & \textbf{0.050}$\pm$0.014  & \textbf{0.042}$\pm$0.008 \\ \cline{2-7}
                             & \multirow{6}{*}{2}       & Vanilla      & 0.675$\pm$0.014 & 0.644$\pm$0.023 & 0.141$\pm$0.003 & 0.143$\pm$0.007 \\ \cline{3-7}
                             &                          & ARL      & 0.663$\pm$0.001 & 0.598$\pm$0.036 & 0.165$\pm$0.015 & 0.171$\pm$0.026 \\
                             &                          & KSMOTE      & 0.628$\pm$0.010  & 0.615$\pm$0.020  & 0.150$\pm$0.012  & 0.148$\pm$0.013 \\
                             &                          & FairRF & 0.631$\pm$0.007 & 0.603$\pm$0.012 & 0.103$\pm$0.019 & 0.105$\pm$0.022 \\ \cline{3-7}
                             &                          & {\m}            & 0.611$\pm$0.021 & 0.658$\pm$0.006 & \textbf{0.085}$\pm$0.022 & \textbf{0.096}$\pm$0.012 \\ \bottomrule
\end{tabularx} \label{table:overall_compas}
\end{table*}
\endgroup

\subsection{Fairness Performances (RQ1)}
To answer \textbf{RQ1}, we compare the proposed framework with the aforementioned baselines in terms of prediction and fairness. All experiments are conducted 5 times and the average performances and standard deviations are reported in  Table~\ref{table:overall_compas}, Table~\ref{table:overall_adult} and Table~\ref{table:text_img} in terms of ACC, F1, $\Delta_{EO}$, $\Delta_{DP}$. We have the following observations:

\begin{itemize}
    \item Adapting sensitive attribute information from a similar source domain could help achieve a better fairness performance in the target domain. For example, compared with ARL, KSMOTE and FairRF, which leverage the implicit sensitive information within the target domain, overall {\m} performs better in terms of both fairness metrics $\Delta_{DP}$ and $\Delta_{EO}$ across all six experiments. 

    \item In general, the proposed framework {\m} can achieve a better performance of fairness, while remains comparatively good prediction performance. For example, in Exp. 6 of CelebA dataset, the performance of $\Delta_{DP}$ of {\m} increases 21.2\% compared to Vanilla, on which no explicit fairness regularization applied, and the performance of ACC only drops 0.1\%.

    \item We conduct experiments on various source and target datasets. We observe that no matter whether the number of instances in the source domain is greater or smaller than the target one, the proposed {\m} outperforms baselines consistently, which demonstrates the intuition of sensitive adaptation is effective and the proposed {\m} is robust.

\end{itemize}

\begingroup
\renewcommand{\arraystretch}{1.3}
\begin{table*}
\centering \caption{Fairness performance comparison in the ADULT dataset. }
\begin{tabularx}{0.89\linewidth}{ccl|llll}
\toprule
\multicolumn{1}{l}{Datasets} & \multicolumn{1}{l}{Exp.} & Methods  & ACC         & F1          & $\Delta_{DP}$          & $\Delta_{EO}$          \\ \hline
\multirow{12}{*}{ADULT}      & \multirow{6}{*}{3}       & Vanilla  & 0.857$\pm$0.003 & 0.655$\pm$0.009 & 0.100$\pm$0.005   & 0.099$\pm$0.014 \\ \cline{3-7}
                             &                          & ARL      & 0.837$\pm$0.006 & 0.668$\pm$0.005 & 0.142$\pm$0.019 & 0.122$\pm$0.016 \\
                             &                          & KSMOTE   & 0.840$\pm$0.002  & 0.511$\pm$0.010  & \textbf{0.072}$\pm$0.008 & 0.138$\pm$0.055 \\
                             &                          & FairRF   & 0.848$\pm$0.003 & 0.594$\pm$0.005 & 0.083$\pm$0.001 & 0.092$\pm$0.015 \\ \cline{3-7}
                             &                          & {\m}   & 0.845$\pm$0.011 & 0.604$\pm$0.008 & 0.080$\pm$0.001 & \textbf{0.073}$\pm$0.016 \\ \cline{2-7}
                             & \multirow{6}{*}{4}       & Vanilla  & 0.839$\pm$0.001 & 0.658$\pm$0.001 & 0.107$\pm$0.003 & 0.093$\pm$0.009 \\ \cline{3-7}
                             &                          & ARL      & 0.816$\pm$0.015 & 0.673$\pm$0.004 & 0.142$\pm$0.025 & 0.079$\pm$0.023 \\
                             &                          & KSMOTE   & 0.824$\pm$0.004 & 0.589$\pm$0.040  & 0.102$\pm$0.016 & 0.140$\pm$0.017  \\
                             &                          & FairRF   & 0.829$\pm$0.003 & 0.611$\pm$0.011 & 0.087$\pm$0.006 & \textbf{0.068}$\pm$0.013 \\ \cline{3-7}
                             &                          & {\m}   & 0.825$\pm$0.004 & 0.599$\pm$0.017 & \textbf{0.085}$\pm$0.004 & 0.070$\pm$0.003  \\ \bottomrule
\end{tabularx} \label{table:overall_adult}
\end{table*}
\endgroup

\begingroup
\renewcommand{\arraystretch}{1.3}

\begin{table*}
\centering \caption{Fairness performance comparison in the Toxicity and CelebA datasets. }
\begin{tabularx}{0.89\linewidth}{ccl|llll}
\toprule
\multicolumn{1}{l}{Datasets} & \multicolumn{1}{l}{Exp.} & Methods  & ACC         & F1          & $\Delta_{DP}$          & $\Delta_{EO}$          \\ \hline
\multirow{6}{*}{Toxicity}    & \multirow{6}{*}{5}       & Vanilla  & 0.761$\pm$0.002 & 0.550$\pm$0.023  & 0.108$\pm$0.017 & 0.118$\pm$0.029 \\ \cline{3-7}
                             &                          & ARL      & 0.759$\pm$0.002 & 0.558$\pm$0.022 & 0.113$\pm$0.016 & 0.115$\pm$0.025 \\
                             &                          & KSMOTE   & 0.619$\pm$0.116 & 0.428$\pm$0.040  & 0.104$\pm$0.028 & 0.095$\pm$0.013 \\
                             &                          & FairRF   & 0.705$\pm$0.024 & 0.475$\pm$0.097 & 0.099$\pm$0.032 & 0.113$\pm$0.006 \\ \cline{3-7}
                             &                          & {\m}   & 0.739$\pm$0.008 & 0.501$\pm$0.011 & \textbf{0.085}$\pm$0.012 & \textbf{0.069}$\pm$0.048 \\ \hline
\multirow{6}{*}{CelebA}      & \multirow{6}{*}{6}       & Vanilla  & 0.864$\pm$0.001 & 0.443$\pm$0.014 & 0.132$\pm$0.010  & 0.237$\pm$0.014 \\ \cline{3-7}
                             &                          & ARL      & 0.801$\pm$0.009 & 0.485$\pm$0.009 & 0.291$\pm$0.012 & 0.359$\pm$0.020  \\
                             &                          & KSMOTE   & 0.859$\pm$0.000     & 0.359$\pm$0.038 & 0.124$\pm$0.020  & 0.226$\pm$0.030   \\
                             &                          & FairRF   & 0.864$\pm$0.001 & 0.405$\pm$0.024 & 0.115$\pm$0.016 & 0.213$\pm$0.026 \\ \cline{3-7}
                             &                          & {\m}   & 0.863$\pm$0.003 & 0.390$\pm$0.030   & \textbf{0.104}$\pm$0.007 & \textbf{0.189}$\pm$0.017 \\ \bottomrule
\end{tabularx} \label{table:text_img}
\end{table*}
\endgroup

\subsection{Ablation Study (RQ2)}
In this section, we aim to analyse the effectiveness of each component in the proposed {\m} framework. 
As shown in Section~\ref{sec:model},  {\m} contains two adversarial components. The first one is designed for making better predictions on target domain's sensitive attribute with the knowledge learned from the source domain. The second one is designed for debiasing with the sensitive attribute predicted by the domain adaptive classifier. In order to answer \textbf{RQ2}, we investigate the effects of these components by defining two variants of {\m}:
\begin{itemize}
    \item \textit{{\m} w/o DA}: This is a variant of {\m} without domain adversary on the source domain. It first trains a classifier with the source domain's sensitive attribute and directly applies it to predict the target domain's sensitive attribute. 
    
    \item \textit{{\m} w/o Debiasing}: This is a variant of {\m} without debiasing adversary on the target domain. In short, it is equivalent to Vanilla. 
    
\end{itemize}



Table~\ref{table:ablation} reports the average performances with standard deviations for each method. We could make the following observations: 
\begin{itemize}
    \item When we eliminate the domain adversary for adapting the sensitive attributes information(i.e., {\m} w/o DA), both $\Delta_{DP}$ and $\Delta_{EO}$ increase, which means the fairness performances are reduced. In addition, the classification performances also decline compared with {\m}. This suggests that the adaptation is able to make better estimations for the target sensitive attributes, which contributes to better fairness performances.
    
    \item When we eliminate the debiasing adversary(i.e., {\m} w/o Debiasing), there is no fairness regularization applied on the target domain. Therefore, it is equivalent to Vanilla. We could observe that the fairness performances drop significantly, which suggests that adding the debiasing adversary to {\m} is indispensable. In addition, even with the adversarial debiasing, {\m} can still achieve comparatively good F1 compared with {\m} w/o Debiasing. This enhances the intuition of leveraging adversary to debias.    
\end{itemize}

{\centering
\begin{table*}[t!]
\caption{Assessing the effectiveness of each component in {\m}.}
\label{table:ablation}
    \centering
    \begin{tabularx}{0.82\textwidth}{l|XXXX}
    \toprule
    Variants &  ACC & F1 & $\Delta_{DP}$ &  $\Delta_{EO}$ \\
    \hline
    {\m} w/o DA     & 0.660$\pm$0.025 & 0.624$\pm$0.042 & 0.062$\pm$0.014 & 0.077$\pm$0.007\\
    {\m} w/o Debiasing  & 0.677$\pm$0.015 & 0.639$\pm$0.018 & 0.128$\pm$0.015 & 0.137$\pm$0.005\\
    \hline
    \textbf{{\m}}      & 0.661$\pm$0.023 & 0.639$\pm$0.016 & 0.050$\pm$0.014 & 0.042$\pm$0.008\\
    \bottomrule
    \end{tabularx}
\end{table*}
}

\begin{subfigure}
\centering
\begin{minipage}{0.6\textwidth}
    \noindent\begin{minipage}[b]{5cm}
        \setcounter{figure}{3}
        \setcounter{subfigure}{0}
        \includegraphics[width=\linewidth]{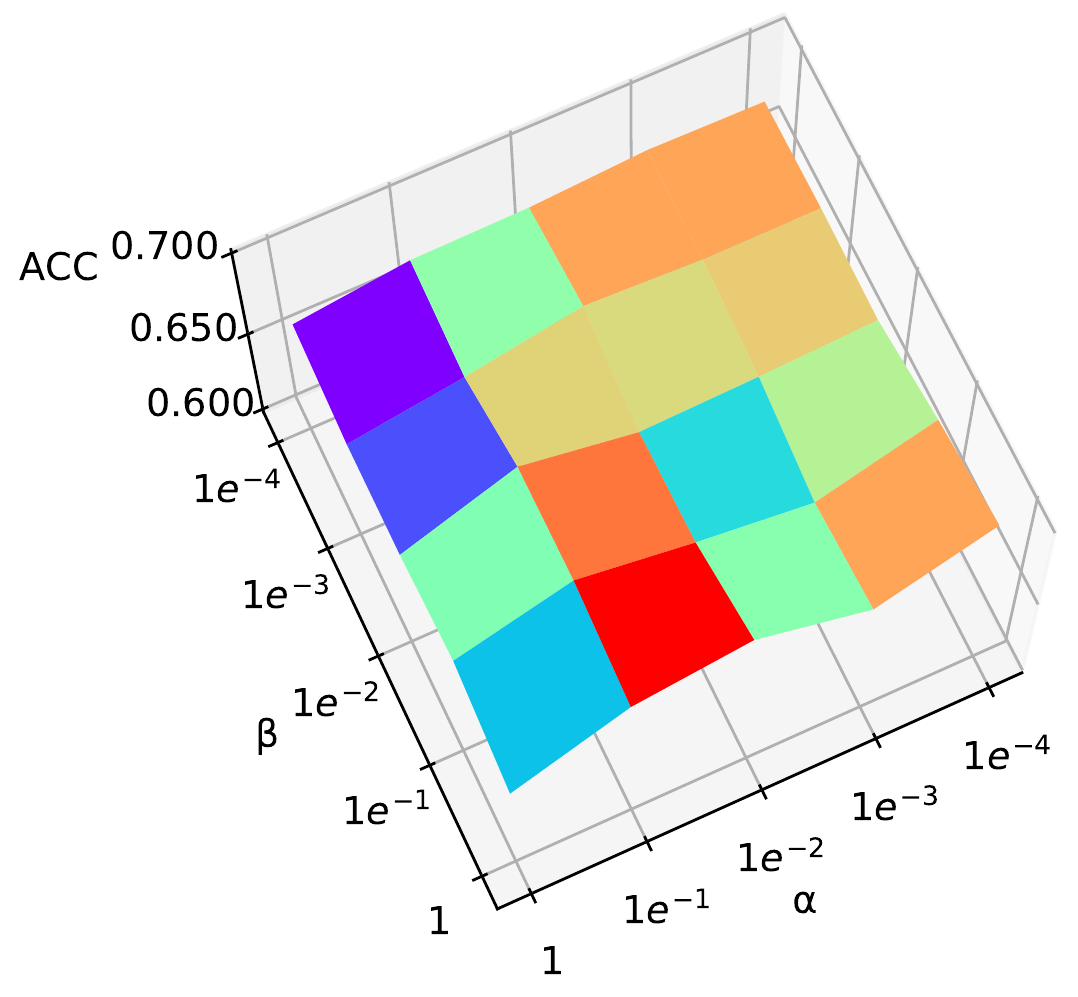}
        \caption{ACC.}
        \label{fig:parameter_acc}
    \end{minipage}
    \hfill
    \begin{minipage}[b]{5cm}
        \setcounter{subfigure}{1}
        \includegraphics[width=\linewidth]{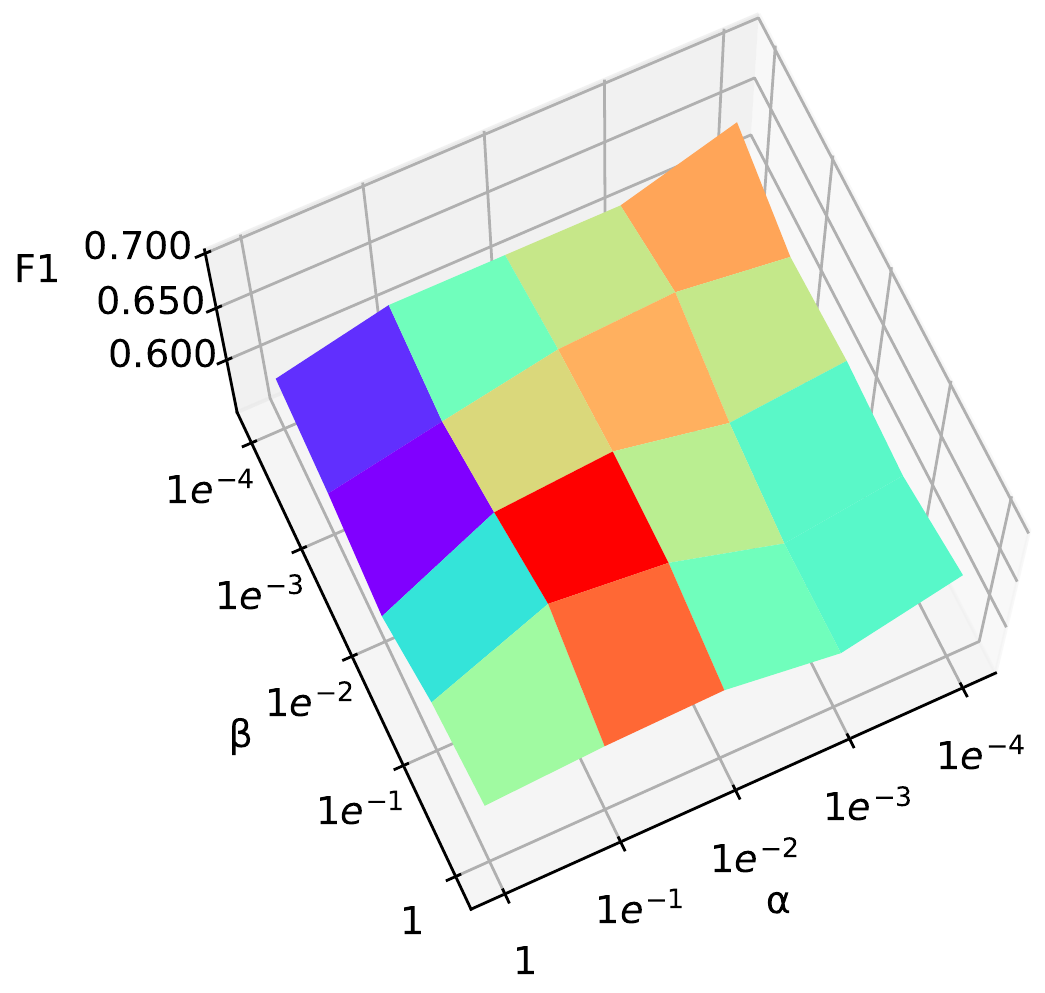}
        \caption{F1.}
        \label{fig:parameter_f1}
    \end{minipage}

    \noindent\begin{minipage}[b]{5cm}
        \setcounter{subfigure}{2}
        \includegraphics[width=\linewidth]{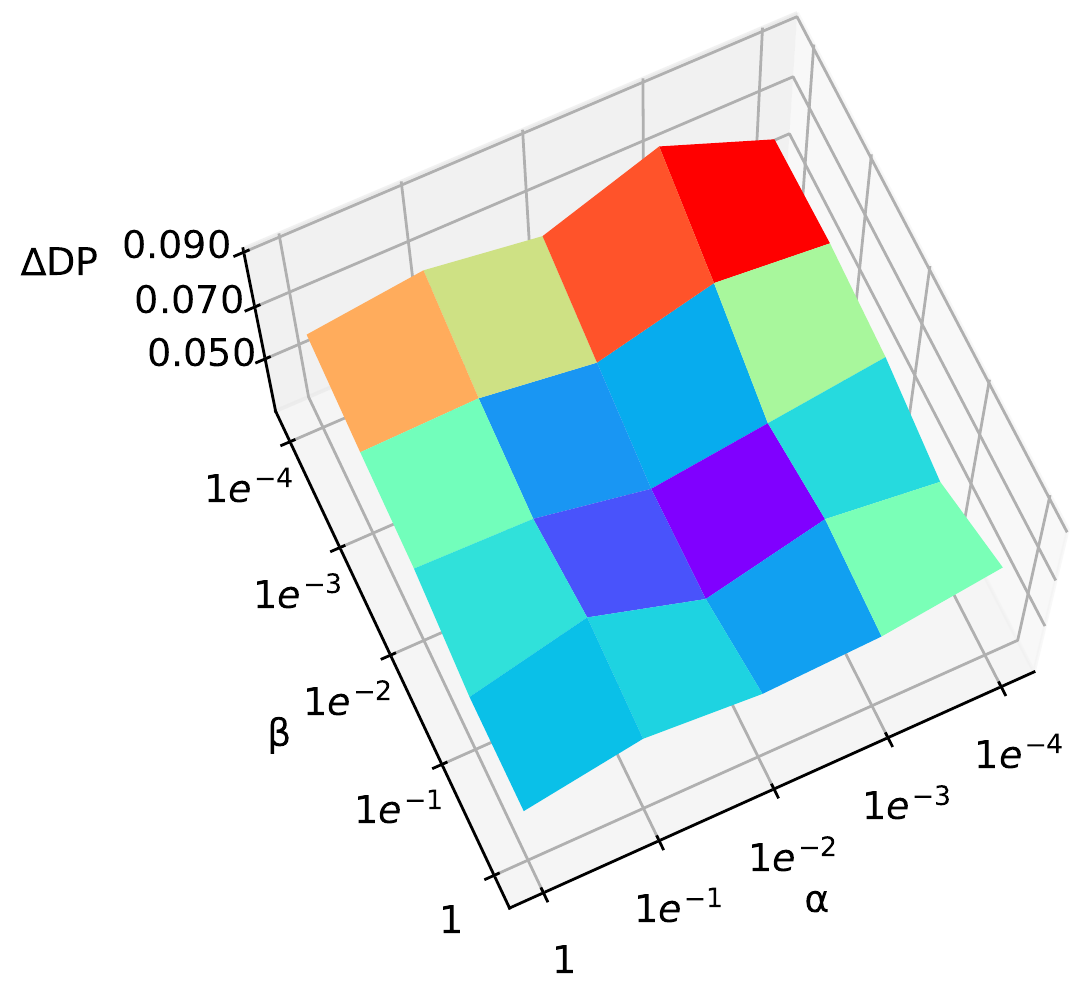}
        \caption{$\Delta_{DP}$.}
        \label{fig:parameter_dp}
    \end{minipage}
    \hfill
    \begin{minipage}[b]{5cm}
        \setcounter{subfigure}{3}
        \includegraphics[width=\linewidth]{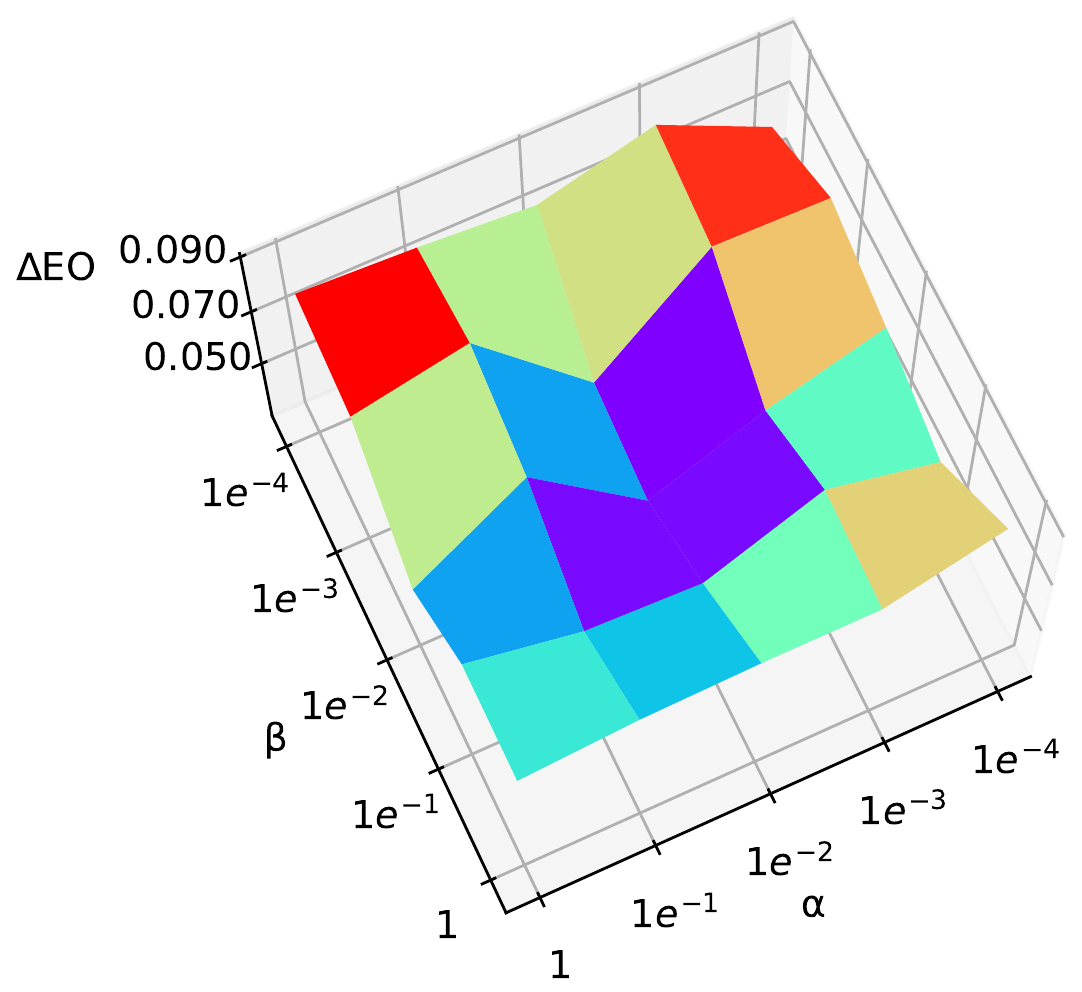}
        \caption{$\Delta_{EO}$.}
        \label{fig:parameter_eo}
    \end{minipage}

\end{minipage}
\setcounter{figure}{3}
\setcounter{subfigure}{-1}
    \caption{Parameter sensitivity analysis to assess the impact of domain adversary and debiasing adversary components.}
    \label{fig:parameter}

\end{subfigure}

\subsection{Parameter Analysis (RQ3)}

To answer \textbf{RQ3}, we explore the parameter sensitivity of the two important hyperparameters of our model, using Exp.1. $\alpha$ controls the impact of the adversary to the sensitive attribute predictor, while $\beta$ controls the influence of the adversary to the debiasing. We vary both $\alpha$ and $\beta$ from [0.0001, 0.001, 0.01, 0.1, 1], and other settings are the same as Exp.1. The results are shown in Fig.\ref{fig:parameter}. 
It is worth noting that, for Fig.\ref{fig:parameter_dp} and Fig.\ref{fig:parameter_eo}, lower values are better fairness performances, while for Fig.\ref{fig:parameter_f1} and Fig.\ref{fig:parameter_acc}, higher values indicate better prediction performance. From Fig.\ref{fig:parameter_dp} and Fig.\ref{fig:parameter_eo} we could observe that: (1) generally, larger $\alpha$ and $\beta$ will achieve fairer predictions, while smaller $\alpha$ and $\beta$ result in worse fairness; (2) when we increase the value of $\alpha$, both $\Delta_{DP}$ and $\Delta_{EO}$ will first decrease and then increase when the value of $\alpha$ is too large. 
This may be because the estimated target sensitive attributes are still noisy and strengthening the adversarial debiasing that optimizes the noisy sensitive attributes may not necessarily lead to better fairness performances; and (3) for $\beta$, the fairness performance is consistently better with a higher value. From Fig.~\ref{fig:parameter_f1}, we observe that the classification performance is better when $\alpha$ and $\beta$ are balanced. Overall, we observe that when $\alpha$ and $\beta$ are in the range of $[0.001, 0.1]$, {\m} can achieve relatively good performances of fairness and classification.

\section{Conclusion and Future Work}\label{sec:conclusion}
In this paper, we study a novel and challenging problem of exploiting domain adaptation for fair and accurate classification for a target domain without the availability of sensitive attributes. We propose a new framework {\m} using a dual adversarial learning approach to achieve a fair and accurate classification. We provide a theoretical analysis to demonstrate that we can learn a fair model prediction under mild assumptions. Experiments on real-world datasets show that the proposed approach can achieve a more fair performance compared to existing approaches by exploiting the information from a source domain, even without knowing the sensitive attributes in the target domain.
For future work, first, we can consider multiple source domains available and explore how to exploit domain discrepancies across multiple domains to enhance the performance of the fair classifier in the target domain. Second, we will explore the fairness transfer under different types of domain shifts, such as the conditional shift of sensitive attributes. Third, we will explore other domain adaption approaches such as meta-transfer learning to achieve cross-domain fair classification.

\bibliographystyle{Frontiers-Harvard}
\bibliography{frontiers}

\end{document}